\documentclass[a4paper]{article}

\usepackage[T1]{fontenc}
\usepackage{amssymb,amsmath,color,amsthm,graphicx}
\usepackage{dsfont,hyperref}
\usepackage{microtype}
\usepackage{graphicx}
\usepackage{subfigure}
\usepackage{booktabs}
\numberwithin{equation}{section}
\newtheorem{thm}{Theorem}[section]
\newtheorem{prop}[thm]{Proposition}
\newtheorem{lemma}[thm]{Lemma}

\newtheorem{asm}{Assumption}[section]

\allowdisplaybreaks

\DeclareMathOperator*{\argmin}{argmin}

\title{Meta-strategy for Learning Tuning Parameters with Guarantees}

\author{Dimitri Meunier (IIT) and Pierre Alquier (RIKEN AIP)}
\date{\today}

\begin{document}

\maketitle

\begin{abstract}
Online learning methods, like the online gradient algorithm (OGA) and exponentially weighted aggregation (EWA), often depend on tuning parameters that are difficult to set in practice. We consider an online meta-learning scenario, and we propose a meta-strategy to learn these parameters from past tasks. Our strategy is based on the minimization of a regret bound. It allows to learn the initialization and the step size in OGA with guarantees. It also allows to learn the prior or the learning rate in EWA. We provide a regret analysis of the strategy. It allows to identify settings where meta-learning indeed improves on learning each task in isolation.
\end{abstract}

\section{Introduction}

In many applications of modern supervised learning, such as medical imaging or robotics, a large number of tasks is available but many of them are associated with a small amount of data. With few datapoints per task, learning them in isolation would give poor results. In this paper, we consider the problem of learning from a (large) sequence of regression or classification tasks with small sample size. By exploiting their similarities we seek to design algorithms that can utilize previous experience to rapidly learn new skills or adapt to new environments.

Inspired by human ingenuity in solving new problems by leveraging prior experience, \emph{meta-learning} is a subfield of machine learning whose goal is to automatically adapt a learning mechanism from past experiences to rapidly learn new tasks with little available data. Since it "learns the learning mechanism" it is also referred to as \emph{learning-to-learn}~\cite{thrun2012learning}. It is seen as a critical problem for the future of machine learning \cite{chollet2019measure}. Numerous formulations exist for meta-learning and we focus on the problem of \emph{online meta-learning} where the tasks arrive one at a time and the goal is to efficiently transfer information from the previous tasks to the new ones such that we learn the new tasks as efficiently as possible (this has also been refered to as {\it lifelong learning}). Each task is in turn processed \emph{online}. To sum up, we have a stream of tasks and for each task a stream of observations.

In order to solve online tasks, diverse well-established strategies exist: perceptron, online gradient algorithm, online mirror descent, follow-the-regularized-leader, exponentially weighted aggregation (also refered to as {\it generalized Bayes} etc. We refer the reader to~\cite{cesa2006prediction,shalev2012online,hazan2016oco,orabona2019modern} for introductions to these algorithms and to so-called regret bounds, that control their generalization errors. We refer to these algorithms as the \emph{within-task} strategies. The big challenge is to design a meta-strategy that uses past experiences to adapt a within-task strategy to perform better on the next tasks.

In this paper we propose a new meta-learning strategy. The main idea to learn the tuning parameters is to minimize its regret bound. We provide a meta-regret analysis for our strategy. We illustrate our results in the case where the within-task strategy is the online gradient algorithm (OGA), and exponentially weighted aggregation (EWA). In the case of OGA, the tuning parameters considered are the initialization and the gradient step. For EWA, we consider either the learning rate, or the prior. In each case, we compare the regret incurred when learning the tasks in isolation to our meta-regret bound. This allows to identify settings where meta-learning indeed improves on learning in isolation.

\subsection{Related works}

Meta-learning is similar to multitask learning~\cite{maurer2006,romera2013multilinear,yamada2017localized} in the sense that the learner faces many tasks to solve. However, in multitask learning, the learner is given a fixed number of tasks, and can learn the connections between these tasks. In meta-learning, the learner must prepare to face future tasks that are not given yet.

Meta-learning is often refered to as learning-to-learn or lifelong learning. \cite{pmlr-v54-alquier17a} proposed the following distinction: ``learning-to-learn'' for situations where the tasks are presented simultaneously, and ``lifelong learning'' for situations where they are presented sequentially. Following this terminology, learning-to-learn algorithms were proposed very early in the literature, with generalization guarantees~\cite{baxter1998theoretical,pentina2014pac,MPR2016,amit2018meta,rothfuss2020pacoh,jose2020transfer}.

On the other hand, in the lifelong learning scenario, until recently, algorithms
were proposed without generalization
guarantees~\cite{ruvolo2013ella,andrychowicz2016learning}. A theoretical study
was proposed by~\cite{pmlr-v54-alquier17a}, but the strategies in this paper are
not feasible in practice. This problem was improved
recently~\cite{denevi2018learning,balcan2019provable,denevi2019learning,finn2019online,zhou2019efficient,fallah2020convergence,denevi2020advantage,konobeev2020optimality}.
In a similar context, in ~\cite{denevi2019online}, the
authors propose an efficient strategy to learn the starting point of OGA. However, an application of this strategy to learning the step
size do not show any improvement over learning in isolation~\cite{dimitri}. The closest work to this paper is \cite{khodak2019adaptive} in
which they also suggest a regret bound minimization strategy. This paper indeed provides a meta-regret bound for learning both the initialization and the gradient step. Note however that this paper remains specific to OGA while our work can be
potentially applied to any online learning algorithm. Indeed, we provide another example: the generalized Bayesian algorithm EWA, for which we learn the prior, or the learning rate. To learn the prior is new in the online setting, up to our knowledge. It can be related to works in the batch setting~\cite{pentina2014pac,amit2018meta,rothfuss2020pacoh,jose2020transfer}, but the improvement with respect to learning in isolation is not quantified in these works.

Finally, it is important to note that we focus on the case where the number of tasks $T$ is large, while the sample size $n$ and algorithmic complexity of each task is moderately small. When each task is extremely complex, for example training a deep neural network on a huge dataset, our procedure (as well as the ones discussed above) will become too expansive. Alternative approaches were proposed, based on optimization via multi-armed bandits~\cite{hyperband,shang}.

\subsection{Organization of the paper}

In Section~\ref{section:notations}, we introduce the formalism of meta-learning and the notations that will be used throughout the paper. In Section~\ref{section:meta}, we introduce our meta-learning strategy, and its theoretical analysis. In Section~\ref{section:example}, we provide the details of our method in the case of meta-learning the initialization and the step size in the online gradient algorithm. Based on our theoretical results, we also explicit situations where meta-learning indeed improves on learning the tasks independently. This is confirmed by experiments reported in this section. In Section~\ref{section:example:Bayes}, we provide the details of our method when the algorithm used within tasks is a pseudo-Bayesian algorithm: EWA. We show how our meta-strategy can be used to tune the learning rate, we also discuss how it can be used to learn priors. The proofs of the main results are given in Section~\ref{section:proofs}.

\section{Notations and preliminaries}
\label{section:notations}

By convention, vectors $v\in\mathbb{R}^d$ are seen as $d\times 1$ matrices (columns). Let $\|v\|$ denote the Euclidean norm of $v$. Let $A^T$ denote the transpose of any $d\times k$ matrix $A$. For two real numbers $a$ and $b$, let $a\vee b = \max(a,b)$ and $a\wedge b =\min(a,b)$. For $z\in\mathbb{R}$, $z_+$ is its positive part $z_+=z\vee 0$.

The learner has to solve tasks $t=1,\dots, T$ sequentially. Each task $t$ consists in $n$ rounds $i=1,\dots,n$. At each round $i$ of task $t$, the learner has to take a decision $\theta_{t,i}$ in a decision space $\Theta\subseteq \mathbb{R}^d$ for some $d>0$. Then, a convex loss function $\ell_{t,i}:\Theta\rightarrow \mathbb{R}$ is revealed to the learner, who incurs the loss $\ell_{t,i}(\theta_{t,i})$. Classical examples with $\Theta\subset\mathbb{R}^d$ include regression tasks, where $\ell_{t,i}(\theta)=(y_{t,i} - x_{t,i}^T \theta)^2$ for some $x_{t,i}\in\mathbb{R}^d$ and $y_{t,i}\in\mathbb{R}$. For classification tasks, $\ell_{t,i}(\theta)=(1-y_{t,i} x_{t,i}^T\theta)_+ $ for some $x_{t,i}\in\mathbb{R}^d$, $y_{t,i}\in\{-1,+1\}$.

\begin{figure*}
\centering
\includegraphics[width=12cm]{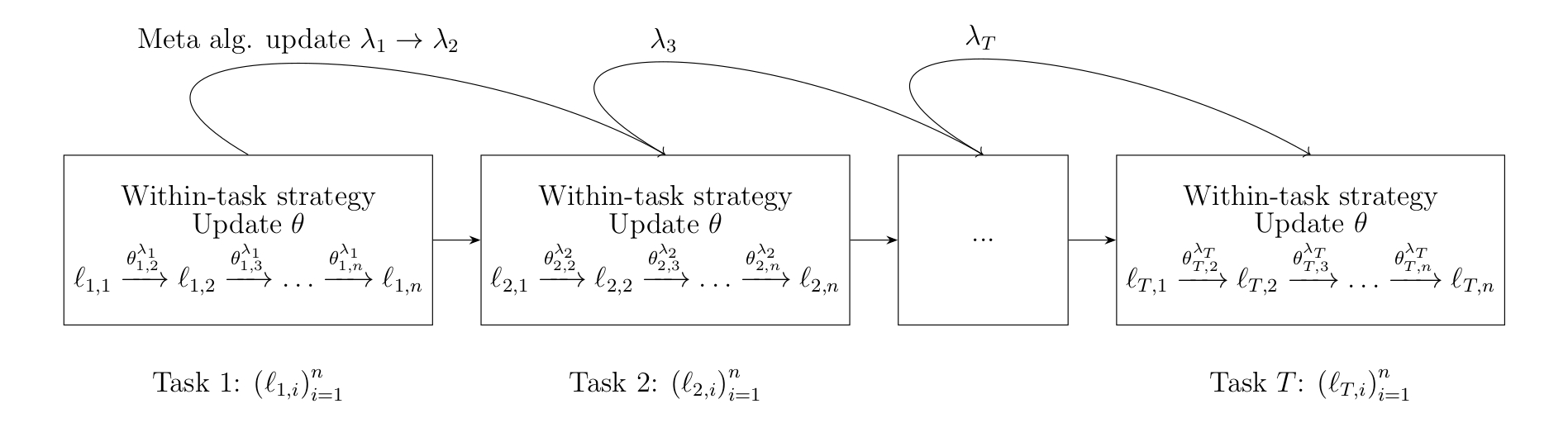}
 \caption{The dynamics of meta-learning.}
\label{graphe_meta_dimitri}
\end{figure*}

Throughout the paper, we will assume that the learner uses for each task an online decision strategy called {\it within-task strategy}, parametrized by a tuning parameter $\lambda\in\Lambda$ where $\Lambda$ is a closed, convex subset of $\mathbb{R}^p$ for some $p>0$. Example of such strategies include the online gradient algorithm (OGA), given by $\theta_{t,i} =\theta_{t,i-1} - \gamma \nabla \ell_{t,i}(\theta_{t,i-1}) $. In this case, the tuning parameters are the initialization, or starting point, $\theta_{t,1}=\vartheta$ and the learning rate, or step size, $\gamma$. That is, $\lambda=(\vartheta,\gamma)$, so $p=d+1$. The parameter $\lambda$ is kept fixed during the whole task. It is of course possible to use the same parameter $\lambda$ in {\it all} the tasks. However, we will be interested here in defining {\it meta-strategies} that will allow to improve $\lambda$ task after task, based on the information available so far. In Section~\ref{section:meta}, we will define such strategies. For now, let $\lambda_t$ denote the tuning parameter used by the learner all along task $t$. Figure~\ref{graphe_meta_dimitri} provides a recap of all the notations.

Let $\theta_{t,i}^\lambda$ denote the decision at round $i$ of task $t$ when the online strategy is used with parameter $\lambda$. We will assume that a regret bound is available for the within-task strategy. By this, we mean that there is a set $\Theta_0\subset\Theta$ of parameters of interest, and that the learner knows a function $\mathcal{B}_n:\Theta\times \Lambda \rightarrow \mathbb{R}$ such that, for any task $t$, for any $\lambda\in\Lambda$,
\begin{equation}
 \sum_{i=1}^n \ell_{t,i}(\theta_{t,i}^\lambda) \leq \underbrace{\inf_{\theta\in\Theta_0} \left\{ \sum_{i=1}^n \ell_{t,i}(\theta)  + \mathcal{B}_n(\theta,\lambda) \right\}}_{=: \mathcal{L}_t(\lambda) }.
\end{equation}
For OGA, regret bounds can be found for example in~\cite{shalev2012online,hazan2016oco} (in this case, $\Theta_0=\Theta$). Other examples include exponentially weighted aggregation (EWA, bounds in~\cite{cesa2006prediction}, here $\Theta_0$ is a finite set of predictors and while decisions $\Theta$ are probability distributions on $\Theta_0$). More examples will be discussed in the paper. The quantity $\mathcal{B}_n(\theta,\lambda)$ is usually refered to as ``the regret''. We will call $\mathcal{L}_t(\lambda)$ the ``meta-loss'': it will be the criterion minimized by our meta-strategy.

The simplest meta-strategy is learning in isolation. That is, we keep $\lambda_t=\lambda_0\in\Lambda$ for all the tasks. The total loss after task $T$ is then given by:
\begin{equation}
\label{equa_meta_fixe}
 \sum_{t=1}^T \sum_{i=1}^n \ell_{t,i}(\theta_{t,i}^{\lambda_0}) \leq \sum_{t=1}^T \mathcal{L}_t(\lambda_0).
\end{equation}
However, when the learner uses a meta-strategy to improve the tuning parameter at the end of each task, the total loss is given by $ \sum_{t=1}^T \sum_{i=1}^n \ell_{t,i}(\theta_{t,i}^{\lambda_t}) $. We will in this paper investigate strategies with meta-regret bounds, that is, bounds of the form
\begin{equation}
\label{equa:meta:regret}
\sum_{t=1}^T \sum_{i=1}^n \ell_{t,i}(\theta_{t,i}^{\lambda_t}) \leq \inf_{\lambda\in\Lambda} \left\{ \sum_{t=1}^T \mathcal{L}_t(\lambda) + \mathcal{C}_T(\lambda) \right\}.
\end{equation}
Of course, such bounds will be relevant only if the right-hand side of~\eqref{equa:meta:regret} is not larger than the right hand side of~\eqref{equa_meta_fixe}, and is significantly smaller in some favorable settings. We show when this is the case in Section~\ref{section:example}.

\section{Meta-learning algorithms}
\label{section:meta}

In this section, we provide two meta-strategies to update $\lambda$ at the end of each task. The first one is feasible only in the special case where we have an explicit formula for the (sub-)gradient of each $\mathcal{L}_t(\lambda)$. In Section~\ref{section:example}, we provide an example where this is the case. The second meta-strategy can be used without this assumption. In both cases, we provide a regret bound as~\eqref{equa:meta:regret}, under the following condition.

\begin{asm}
 \label{asm:regret}
 For any $t\in\{1,\dots,T\}$, the function $\lambda\mapsto\mathcal{L}_t(\lambda)$ is $L$-Lipschitz and convex.
\end{asm}

\subsection{Special case: the gradient of the meta-loss is available in closed form}

As each $\mathcal{L}_t$ is convex, its subdifferential at each point of $\Lambda$ is non-empty. For the sake of simplicity, we will use the notation $\lambda\mapsto\nabla \mathcal{L}_t(\lambda)$ in the following formulas to denote {\it any} element of its subdifferential at $\lambda$. We define the online gradient meta-strategy (OGMS) with step $\alpha>0$ and starting point $\lambda_1\in\Lambda$: for any $t>1$,
\begin{equation}
 \label{dfn:lambda:OGMS}
 \lambda_{t} = \Pi_{\Lambda}[\lambda_{t-1} - \alpha \nabla \mathcal{L}_{t-1}(\lambda_{t-1})]
\end{equation}
where $\Pi_\Lambda$ denotes the orthogonal projection on $\Lambda$.

\subsection{The general case}

We now cover the general case, where a formula for the gradient of $\mathcal{L}_t(\lambda)$ might not be available. We propose the online proximal meta-strategy (OPMS) with step $\alpha>0$ and starting point $\lambda_1\in\Lambda$, defined by:
\begin{equation}
 \label{dfn:lambda:OPMS}
 \lambda_{t} = \argmin_{\lambda\in\Lambda} \left\{ \mathcal{L}_{t-1}(\lambda) + \frac{\|\lambda - \lambda_{t-1} \|^2}{2\alpha} \right\}.
\end{equation}
Using classical notations, e.g~\cite{parikh2014proximal}, we can rewrite this definition with the proximal operator (hence the name of the method). Indeed
$
 \lambda_{t} = {\rm prox}_{\alpha \mathcal{L}_{t-1}}(\lambda_{t-1})
$
where ${\rm prox}$ is the proximal operator given by, for any $x\in\Lambda$ and any convex function $f:\Lambda\rightarrow\mathbb{R}$,
\begin{equation}
 {\rm prox}_{f}(x) = \argmin_{\lambda\in\Lambda}\left\{ f(\lambda) + \frac{\|x-\lambda\|^2}{2} \right\}.
\end{equation}

This strategy is feasible in practice in the regime we are interested in, that is, when $n$ is small or moderately large, and $T\rightarrow \infty$. The learner has to store all the losses of the current task $\ell_{t-1,1},\dots,\ell_{t-1,n}$. At the end of the task, the learner can use any convex optimization algorithm to minimize, with respect to $(\theta,\lambda)\in\Theta \times \Lambda$, the function
\begin{equation} \label{eq:prox_update}
F_t(\theta,\lambda) = \sum_{i=1}^n \ell_{t,i}(\theta)  + \mathcal{B}_n(\theta,\lambda) + \frac{\|\lambda-\lambda_{t-1}\|^2}{2\alpha}.
\end{equation}
We can use a (projected) gradient descent on $F_t$ or its accelerated
variants~\cite{nesterov2013introductory}. 

\subsection{Regret analysis}

\begin{prop}
\label{prop:OPMS:OGMS}
Under Assumption~\ref{asm:regret}, using either OGMS or OPMS with step $\alpha>0$ and starting point $\lambda_1\in\Lambda$ leads to
\begin{equation}
\sum_{t=1}^T \sum_{i=1}^n \ell_{t,i}(\theta_{t,i}^{\lambda_t})
\leq \inf_{\lambda\in\Lambda} \left\{ \sum_{t=1}^T \mathcal{L}_t(\lambda) + \frac{\alpha TL^2}{2} + \frac{\|\lambda-\lambda_1\|^2}{2\alpha} \right\}.
\end{equation}
\end{prop}

The proof can be found in Section~\ref{section:proofs}.

\section{Example: learning the tuning parameters of online gradient descent}
\label{section:example}

In all this section, we work under the following condition.
\begin{asm}
 \label{asm:loss}
 For any $(t,i)\in\{1,\dots,T\}\times\{1,\dots,n\}$, the function $\ell_{t,i}$ is $\Gamma$-Lipschitz and convex.
\end{asm}

\subsection{Explicit meta-regret bound}

We study the situation where the learner uses (projected) OGA as a within-task strategy, that is $\Theta=\{\theta\in\mathbb{R}^d: \|\theta\|\leq C \}$ and, for any $i>1$,
\begin{equation}
 \theta_{t,i} = \Pi_{\Theta} [ \theta_{t,i-1} - \gamma \nabla \ell_{t,i} (\theta_{t,i-1})].
\end{equation}
With such a strategy, we already mentioned that $\lambda=(\vartheta,\gamma)\in\Lambda \subset \Theta\times \mathbb{R}_+$ contains an initialization and a step size. An application of the results in Chapter 11 in~\cite{cesa2006prediction} gives
$\mathcal{B}_{n}(\theta,\lambda) = \mathcal{B}_{n}(\theta,(\vartheta,\gamma))  = \gamma \Gamma^2 n /2 + \|\theta-\vartheta\|^2/(2\gamma)$. So
\begin{equation}
 \mathcal{L}_t((\vartheta,\gamma)) = 
 \inf_{ \|\theta\|\leq C} \left\{ \sum_{i=1}^n \ell_{t,i}(\theta)  + \frac{\gamma \Gamma^2 n}{2} + \frac{\|\theta-\vartheta\|^2}{2\gamma} \right\}.
\end{equation}
It is quite direct to check Assumption~\eqref{asm:regret}. We summarize this in the following proposition.
\begin{prop}
 \label{prop:asm:OGA}
 Under Assumption~\ref{asm:loss}, assume that the learner uses OGA as an inner algorithm. Assume $\Lambda = \{\vartheta\in\mathbb{R}^d:\|\vartheta\|\leq C\} \times [\underline{\gamma},\bar{\gamma}]$ for some $C>0$ and $0<\underline{\gamma}<\bar{\gamma}<\infty$. Then Assumption~\ref{asm:regret} is satisfied with
 \begin{equation}
 \label{equa:L}
  L := \sqrt{\frac{n^2 \Gamma^4}{4} + \frac{4C^2}{\underline{\gamma}^2} + \frac{4C^4}{\underline{\gamma}^4} }.
 \end{equation}
\end{prop}

So, when the learner uses one of the meta-strategies OGMS or OPMS, we can apply Proposition~\ref{prop:OPMS:OGMS} respectively. This leads to the following theorem.
\begin{thm}
 \label{thm:main}
 Under the assumptions of Proposition~\ref{prop:asm:OGA}, with $\underline{\gamma} = 1/n^\beta$ for some $\beta>0$ and $\bar{\gamma}=C^2$, when the learner uses either OGMS or OPMS with
 \begin{equation}
  \alpha = \frac{C}{L}\sqrt{\frac{4+C^2}{T}}
 \end{equation}
 (where $L$ is given by~\eqref{equa:L}), we have:
 \begin{multline}
 \sum_{t=1}^T \sum_{i=1}^n \ell_{t,i}(\theta_{t,i}^{\lambda_t})
\leq
\inf_{\theta_1,\dots,\theta_T \in \Theta} \Biggl\{
 \sum_{t=1}^T \sum_{i=1}^n \ell_{t,i}(\theta_{t})
 \\
 + \mathcal{C}(\beta,\Gamma,C) \Biggl[
 n^{1\vee 2\beta} \sqrt{T} + \Biggl( n^{1-\beta} + \sigma(\theta_1^T) \sqrt{n} \Biggr) T
 \Biggr]
\Biggr\}
\end{multline}
where $\mathcal{C}(\beta,\Gamma,C)>0$ depends only on $(\beta,\Gamma,C)$ and where:
\begin{equation}
 \sigma(\theta_1^T) = \sqrt{\frac{1}{T}\sum_{t=1}^T \left\| \theta_t - \frac{1}{T}\sum_{s=1}^T \theta_s \right\|^2 }.
\end{equation}
\end{thm}

Let us compare this result with learning in isolation. For a $\gamma$ in $1/\sqrt{n}$, OGA leads to a regret in $\sqrt{n}$. After $T$ tasks, learning in isolation thus leads to a regret in $T\sqrt{n}$. Our strategies with $\beta=1$ lead to a regret in
\begin{equation}
 n^2 \sqrt{T} + \left( 1 + \sigma(\theta_1^T) \sqrt{n} \right) T.
\end{equation}
The term $n^2 \sqrt{T}$ is the price to pay for meta-learning. In the regime we are interested in (small $n$, large $T$), it is smaller than $T\sqrt{n}$. Consider the leading term. In the worst case, it is also in $T\sqrt{n}$. However, when there are good predictors $\theta_1,\dots,\theta_T$ for tasks $1,\dots,T$ respectively such that $ \sigma(\theta_1^T) $ is small, we see the improvement with respect to learning in isolation. The extreme case is when there is a good predictor $\theta^*$ that predicts well for all the tasks. In this case, the regret with respect to $\theta_1=\dots=\theta_T=\theta^*$ is in $ n^2 \sqrt{T} +  T$, which improves significantly on learning in isolation. Note however than, using a different meta-strategy, specifically designed for OGA,~\cite{khodak2019adaptive} obtain a better dependence on $T$ when $\sigma(\theta_1^T)=0$.

Let us now discuss the implementation of our meta-stategy. We first remark that under the quadratic loss, it is possible to derive a formula for $\mathcal{L}_t$, which allows to use OGMS. We then discuss OPMS for the general case.

\subsection{Special case: quadratic loss}

First, consider $\ell_{t,i} = (y_{t,i} - x_{t,i}^T \theta)^2$ for some $y_{t,i}\in\mathbb{R}$ and $x_{t,i}\in\mathbb{R}^d$. Assumption~\ref{asm:loss} is satisfied if we assume moreover that all the $|y_{t,i}|\leq c$ and $\|x_{t,i}\| \leq b$, with $\Gamma=2bc+2b^2C$. In this case,
\begin{equation}
\label{prog:ridge}
 \mathcal{L}_t((\vartheta,\gamma)) =
 \inf_{ \|\theta\|\leq C} \Biggl\{ \sum_{i=1}^n (y_{t,i}-x_{t,i}^T \theta)^2  
 + \frac{\gamma \Gamma^2 n}{2} + \frac{\|\theta-\vartheta\|^2}{2\gamma} \Biggr\}.
\end{equation}
Define $Y_t = (y_{t,1},\dots,y_{t,n})^T$ and $X_{t} = (x_{t,1}|\dots|x_{t,n})^T$. The minimizer of $\sum_{i=1}^n (y_{t,i}-x_{t,i}^T \theta)^2  + \|\theta-\vartheta\|^2/(2\gamma)$ with respect to $\theta$ is known as the ridge regression estimator:
\begin{equation}
 \hat{\theta}_t = \left( X_t^T X_t + \frac{I}{2\gamma}\right)^{-1} \left( X_t^T Y_t + \frac{\vartheta}{2\gamma} \right).
\end{equation}
It also coincides with the minimizer in the right-hand-side of~\eqref{prog:ridge} on the condition that $\|\hat{\theta}_t\|\leq C$. In this case, by pluging $\hat{\theta_t}$ in~\eqref{prog:ridge}, we have a close form formula for $\mathcal{L}_t((\vartheta,\gamma))$, and an explicit (but cumbersome) formula for its gradient. It is thus possible to use the OGMS strategy to update $\lambda=(\vartheta,\gamma)$. 

\subsection{The general case}

In the general case, denote $\lambda_{t-1} = (\vartheta_{t-1},\gamma_{t-1})$, then $\lambda_t = (\vartheta_t,\gamma_t)$ is obtained by minimizing
\begin{multline} \label{eq:opms_obj}
F_t(\theta,  (\vartheta,\gamma)) = \sum_{i=1}^n \ell_{t,i}(\theta) + \frac{\gamma \Gamma^2 n}{2} 
\\ + \frac{\|\theta-\vartheta\|^2}{2\gamma} + \frac{\|\vartheta-\vartheta_{t-1}\|^2 + (\gamma-\gamma_{t-1})^2}{2\alpha}
\end{multline}
with respect to $\theta,\vartheta,\gamma$. Any efficient minimization procedure can be used. In our experiments, we used a projected gradient descent, the gradient being given by:
\begin{align}
 \frac{\partial F_t}{\partial \theta} & = \sum_{i=1}^n \nabla \ell_{t,i}(\theta) + \frac{\theta-\vartheta}{\gamma},
 \\
 \frac{\partial F_t}{\partial \vartheta} & = \frac{\vartheta-\theta}{\gamma} + \frac{\vartheta-\vartheta_{t-1}}{\alpha},
 \\
  \frac{\partial F_t}{\partial \gamma} & = \frac{\Gamma^2 n}{2} - \frac{\|\theta-\vartheta\|^2}{2\gamma^2} + \frac{\gamma-\gamma_{t-1}}{\alpha}.
\end{align}
Note that even though we do not {\it stricto sensu} obtain the minimizer of
$F_t$, we can get arbitrarily close to it by taking a large enough number of
steps. The main difference between this algorithm and the
  strategy suggested in~\cite{khodak2019adaptive} is that it is obtained by
  applying the general proximal update introduced in Equation
  \ref{eq:prox_update}, while they decoupled the update for the initialization step and the
  learning rate.

\subsection{Experimental study}
In this section we compare on simulated data the numerical performance of OPMS w.r.t learning the task in isolation with online gradient descent (I-OGA).  To measure the impact of learning the gradient step $\gamma$,  we also introduce mean-OPMS that uses the same strategy as OPMS but only learns the starting point $\vartheta$ (it is thus close to~\cite{denevi2019online}). We present the results for regression tasks with the mean-squared-error loss, and then for classification with the hinge loss. The notebooks of the experiments can be found online:~\url{https://dimitri-meunier.github.io/}.

\begin{figure*}
\centering
\includegraphics[width=12cm]{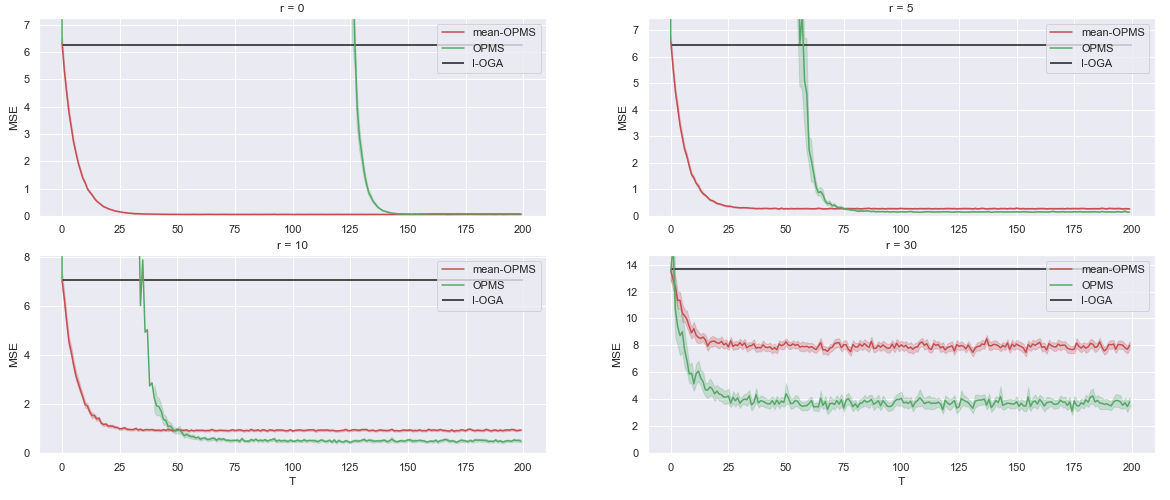}
 \caption{Performance of learning in isolation with OGA ({\bf I-OGA}), OPMS to learn the initialization ({\bf mean-OPMS}) and OPMS to learn the initialization and step size ({\bf OPMS}). We report the average end-of-task MSE losses at the end of each task, for different values of the task-similarity index $r \in \{0,5,10,30\}$. The results are averaged over 50 independent runs to get confidence intervals.}
 \label{fig_regression}
\end{figure*}

\paragraph{Synthetic Regression.} At each round $t = 1, \ldots, T$,  the meta learner receives sequentially a regression task that corresponds to a dataset $(x_{t,i}, y_{t,i})_{i=1,\ldots,n}$ generated as $y_{t,i} = x_{t,i}^T \theta_t + \epsilon_{t,i}$, $x_{t,i} \in \mathbb{R}^d$. The noise is $\epsilon_{t,i} \sim \mathcal{U}([-\sigma^2,\sigma^2])$,  the inputs are uniformly sampled on the $(d-1)$-unit sphere $\mathcal{S}^{d-1}$  and $\theta_t = r u  + \theta_0$, $u \sim \mathcal{U}\left(\mathcal{S}^{d-1} \right)$, $\theta_0 \in \mathbb{R}^d$,  $r \in \mathbb{R}_+$.  We take $d=20$, $n=30$, $T=200$,  $\sigma^2=0.5$ and $\theta_0$ with all components equal to $5$. In this setting, $\theta_0$ is a common bias between the tasks,  $\sigma^2$ is the inter-task variance and $r$ characterizes the tasks similarity. We experiment with different values of $r \in \{0,5,10,30\}$ to observe the impact of task similarity on the meta-learning process. The smaller $r$, the closer are the tasks and for the extreme case of $r=0$ the tasks are identical. We draw attention to the fact that a cross-validation procedure to select $\alpha$ or $\lambda$ is not valid in the online setting as it would require to know several tasks in advance for the former and several datapoints in advance for each task for the latter. Moreover, the theoretical values are based on worst-case analysis and lead in practice to slow learning. In practice, to set these values to the correct order of magnitude without adjusting the constants led to better results. So, for mean-OPMS and OPMS we set $\alpha = 1/\sqrt{T}$, for OPMS and I-OGA we set $\lambda = 1/\sqrt{n}$. Instead of cross-validation, one can launch several online learners in parallel with different parameters values to pick the best one (or aggregate them). That is the strategy we use to select $\Gamma$ for OPMS. Note that the exact value of $\Gamma$ is usually unkown in practice; its automatic calibration is an important open question. To solve (\ref{eq:opms_obj}), after each task we use the exact solution for mean-OPMS and projected Newton descent with 10 steps for OPMS. We observed that not reaching the exact solution of (\ref{eq:opms_obj}) does not harm the performance of the algorithm and 10 steps are sufficient to reach convergence. The results are displayed in Table \ref{table:results_regression} and Figure \ref{fig_regression}.  On Figure \ref{fig_regression},  for each task $t=1,\dots,T$, we report the average end-of-task loss $MSE_t=\sum_{i=1}^n \ell_{t,i}(\theta_{t,n})/n$ averaged over 50 independent runs (with their confidence intervals).  Table \ref{table:results_regression} reports $MSE_t$ averaged over the 100 last tasks. The results confirms our theoretical findings: learning $\gamma$ can bring a substantial benefit over just learning the starting point, which in turn brings a considerable benefit with respect to learning the tasks in isolation. Learning the gradient step makes the meta-learner more robust to task dissimilarities (i.e. when $r$ increases) as shown in Figure \ref{fig_regression}. In the regime where $r$ is low, learning the gradient step does not help the meta-learner as it takes more steps to reach convergence. Overall both meta learners are consistently better than learning the task in isolation since the number of observation per task is low. 

\begin{table}[h!]
\centering
\begin{tabular}{lrrrr}
\toprule
{} &   r=0 &   r=5 &  r=10 &   r=30 \\
\midrule
I-OGA     &  6.24 &  6.44 &  7.06 &  13.60 \\
mean OPMS &  0.05 &  0.27 &  0.93 &   7.93 \\
OPMS      &  0.07 &  0.15 &  0.49 &   3.72 \\
\bottomrule
\end{tabular}
\caption{Average end-of-task MSE of the 100 last tasks (averaged over 50 independent runs). }
\label{table:results_regression}
\end{table}

\paragraph{Synthetic Classification.}  At each round $t = 1, \ldots, T$,  the meta learner receives sequentially a binary classification task with the Hinge loss that corresponds to a dataset $(x_{t,i}, y_{t,i})_{i=1,\ldots,n}$.  The binary labels $\{-1,1\}$ are generated as a logistic model $\mathbb{P}(y=1) = (1+\exp(-x^t \theta_t))^{-1}$. The task parameters $\theta_t$ and the inputs are generated like in the regression setting.  To add some noise we shuffle $10\%$ of the labels.  We take $d=10$, $n=100$, $T=500$, $r=2$.  For mean-OPMS and OPMS we set $\alpha = 1/\sqrt{T}$, for OPMS and I-OGA we set $\lambda = 1/\sqrt{n}$.  For the optimisation of $F_t$ (\ref{eq:opms_obj}) with both OPMS and mean-OPMS we use a projected gradient descent with $50$ steps.

\begin{figure}[ht]
 \caption{Performance of learning in isolation with OGA ({\bf I-OGA}), OPMS to learn the initialization ({\bf mean-OPMS}) and OPMS to learn the initialization and step size ({\bf OPMS}) on a sequence of classification tasks with the Hinge loss.  We report the meta-regret of the Hinge loss.  The results are averaged over 10 independent runs (dataset generation) to get confidence intervals.}
\label{fig_classification}
\centering
\includegraphics[width=8cm]{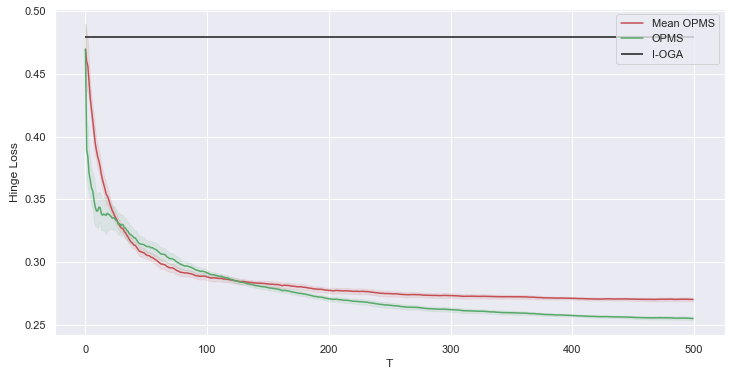}
\end{figure}

On Figure \ref{fig_classification},  for each task $t=1,\dots,T$, we report the regret on the end-of-task losses: $R(t)=\frac{1}{nt} \sum_{k=1}^t \sum_{i=1}^n \ell_{k,i}(\theta_{k,n})$, averaged over 10 independent runs (with their confidence intervals).  As the for regression setting, the results confirm our theoretical findings: by learning $\gamma$ (OPMS) we reach a better overall performance than just learning the initialization (mean-OPMS) and both a substantially stronger than independent task learning (I-OGA).  Let us note that in the classification regime there is no known closed formed expression for the meta-gradient, therefore OGMS cannot be used.

\section{Second example: learning the prior or the learning rate in exponentially weighted aggregation}
\label{section:example:Bayes}

In this section, we will study a generalized Bayesian method, exponentially weighted aggregation (EWA). Consider a {\it finite} set $\Theta_0 = \{\theta_1,\dots,\theta_M\} \subset \mathbb{R}^d$. EWA depends on a prior distribution $\pi$ on $\Theta_0$, and on a learning rate $\eta>0$, and returns a decision in $\Theta={\rm conv}(\theta_1,\dots,\theta_M)$ the convex enveloppe of $\Theta_0$. In all this section, we work under the following condition.
\begin{asm}
 \label{asm:loss:Bayes}
 There is a $B\in\mathbb{R}_+^*$ such that for any $(t,i)\in\{1,\dots,T\}\times\{1,\dots,n\}$, the function $\ell_{t,i}$ is $\Theta \rightarrow [0,B]$ and convex.
\end{asm}
We will sometimes use a stronger assumption.
\begin{asm}
 \label{asm:loss:Bayes:2}
 There is a $C\in\mathbb{R}_+^*$ such that for any $(t,i)\in\{1,\dots,T\}\times\{1,\dots,n\}$, the function $\theta\mapsto  \exp(-\ell_{t,i}(\theta)/C)$ is concave.
\end{asm}
Examples of situation where Assumption~\ref{asm:loss:Bayes:2} is satisfied are provided in~\cite{cesa2006prediction}. Note that Assumption~\ref{asm:loss:Bayes:2} implies Assumption~\ref{asm:loss:Bayes}.

\subsection{Reminder on EWA}

The update in EWA is given by:
\begin{equation}
 \theta_{t,i} = \sum_{\theta\in\Theta_0} p_{t,i}(\theta) \theta
\end{equation}
where $p_{t,i}$ are weights defined by
\begin{equation}
 p_{t,i}(\theta) = \frac{\exp\left[-\eta \sum_{j=1}^{i-1} \ell_{t,j}(\theta) \right] \pi(\theta) }
 {\sum_{\vartheta\in\Theta_0} \exp\left[-\eta \sum_{j=1}^{i-1} \ell_{t,j}(\vartheta) \right] \pi(\vartheta)}.
\end{equation}
The strategy is studied in detail in~\cite{cesa2006prediction}. We refer the reader to~\cite{alquier} and the references therein for connections to Bayesian inference. We remind the following regret bounds from~\cite{cesa2006prediction}. First, under Assumption~\ref{asm:loss:Bayes},
\begin{equation}
\label{regret:slow}
 \sum_{i=1}^n \ell_{t,i}(\theta_{t,i}) \leq \min_{\theta\in\Theta_0} \left[ \sum_{i=1}^n \ell_{t,i}(\theta)
 + \frac{\eta n B^2}{8} + \frac{\log\frac{1}{\pi(\theta)}}{\eta}
 \right].
\end{equation}
Moreover, under the stronger Assumption~\ref{asm:loss:Bayes:2},
\begin{equation}
\label{regret:fast}
 \sum_{i=1}^n \ell_{t,i}(\theta_{t,i}) \leq \min_{\theta\in\Theta_0} \left[ \sum_{i=1}^n \ell_{t,i}(\theta)
 + C \log\frac{1}{\pi(\theta)}
 \right].
\end{equation}

In Subsection~\ref{learning:eta}, we work in the general setting (Assumption~\ref{asm:loss:Bayes}), and we use our meta-strategy OPMS or OGMS to learn $\eta$. In Subsection~\ref{learning:pi}, we use OPMS or OGMS to learn $\pi$ under Assumption~\ref{asm:loss:Bayes:2}.

\subsection{Learning the rate $\eta$}
\label{learning:eta}

Consider the uniform prior $\pi(\theta)=1/M$ for any $\theta\in\Theta_0$. Then, the regret bound~\eqref{regret:slow} becomes:
\begin{equation}
 \sum_{i=1}^n \ell_{t,i}(\theta_{t,i}) \leq \min_{\theta\in\Theta_0} \sum_{i=1}^n \ell_{t,i}(\theta)
 + \frac{\eta n B^2}{8} + \frac{\log M}{\eta}
\end{equation}
and it is then possible to optimize it explicitly with respect to $\eta$. The value minimizing the bound is $\eta=(2/B)\sqrt{2\log(M)/n}$ and the regret bound becomes:
\begin{equation}
 \sum_{i=1}^n \ell_{t,i}(\theta_{t,i}) \leq \min_{\theta\in\Theta_0}  \sum_{i=1}^n \ell_{t,i}(\theta)
 + B \sqrt{\frac{n \log M}{2}}.
\end{equation}

In practice, however, while it is often reasonnable to assume that the loss function is bounded (as in Assumption~\ref{asm:loss:Bayes}), very often, one does not know a tight upper bound. Thus, one may use a constant $B$ that satisfies Assumption~\ref{asm:loss:Bayes}, but that is far too large. Even though one does not know a better upper bound than $B$, one would like a regret bound that depends on the tightest possible upper bound.

In the meta-learning framework, define:
\begin{equation}
\label{learning:eta:criterion}
 \mathcal{L}_t(\eta) = \min_{\theta\in\Theta_0} \sum_{i=1}^n \ell_{t,i}(\theta)
 + \frac{\eta n \left[\max_{\vartheta\in\Theta_0,1\leq i\leq n}\ell_{t,i}(\vartheta)\right]^2 }{8} + \frac{\log M}{\eta}
\end{equation}
for $\eta\in \Lambda = [1/n,1]$. It is immediate to prove that this function is convex and $L$-Lipschitz with $L=n^2\log(M)+ nB^2/8$. So Assumption~\ref{asm:regret} is satisfied, allowing to use the OPMS or OGMS strategy without having to know a tight upper bound on the losses. Note that in this context, the OGMS strategy is given by:
$$ \eta_t = \frac{1}{n}\vee \left[ \eta_{t-1} - \alpha\left(  \frac{n \left[\max_{\theta\in\Theta_0,1\leq i\leq n}\ell_{t,i}(\theta)\right]^2 }{8} - \frac{\log M}{\eta_{t-1}^2} \right)\right]\wedge 1. $$

\begin{thm}
 \label{thm:learning:eta}
 Under Assumption~\ref{asm:loss:Bayes}, using OGMS or OPMS on $\mathcal{L}_t(\eta)$ as in~\ref{learning:eta:criterion} with $\eta_1=1$, $L=n^2\log(M)+ nB^2/8$ and
\begin{equation}
 \alpha = \frac{1}{L}\sqrt{\frac{2}{T}}
\end{equation}
 we have
\begin{multline}
   \sum_{t=1}^T \sum_{i=1}^n \ell_{t,i}(\theta_{t,i}^{\eta_t}) \leq  \sum_{t=1}^T  \min_{\theta\in\Theta_0} \sum_{i=1}^n \ell_{t,i}(\theta)
   +  b T \sqrt{\frac{n \log(M)}{2}}
   \\
   + T\log(M)+ \frac{b^2 T }{8}
   + \left(n^2\log M + \frac{nB^2}{8} \right)\sqrt{2T}
\end{multline}
 where
 \begin{equation}
  b = \max_{\theta\in\Theta_0,1\leq t\leq T,1\leq i\leq n}|\ell_{t,i}(\theta)|.
 \end{equation}
\end{thm}

When learning in isolation with $\eta_0 = (2/B)\sqrt{2\log(M)/n}$, the meta-regret is in $B T \sqrt{n \log(M)/2}$. On the other hand, meta-learning leads to a meta-regret in $b T \sqrt{n \log(M)/2} +n^2\log M \sqrt{2T}+o(T\sqrt{n} + n^2\sqrt{T})$. In other words, we replace the potentially loose upper bound $B$ by the tightest possible bound $b$, at the cost of an additional $n^2\log M \sqrt{2T}$ term. Here again, when $T$ is large enough with respect to $n$, this term is negligible.

\subsection{Learning the prior $\pi$}
\label{learning:pi}

Under Assumption~\ref{asm:loss:Bayes:2}, we have the regret bound in~\eqref{regret:fast}. Without any information on $\Theta_0$, it seems natural to use the uniform prior $\pi$ on $\Theta_0=\{\theta_1,\dots,\theta_M\}$, which leads to
\begin{equation}
 \sum_{i=1}^n \ell_{t,i}(\theta_{t,i}) \leq \min_{\theta\in\Theta_0}  \sum_{i=1}^n \ell_{t,i}(\theta)
 + C \log M.
\end{equation}
If some additional information was available, like for example: ``the best $\theta$ is always either $\theta_1$ or $\theta_2$'', one would rather chose the uniform prior on $\{\theta_1,\theta_2\}$, and obtain the bound:
\begin{equation}
 \sum_{i=1}^n \ell_{t,i}(\theta_{t,i}) \leq \min_{\theta\in\Theta_0}  \sum_{i=1}^n \ell_{t,i}(\theta)
 + C \log 2.
\end{equation}
Unfortunately, such an information is generally not available. However, in the context of meta-learning, we can take advantage of the previous tasks to learn such an information.

Thus, let us define, for any task $t$,
\begin{equation}
\label{thetatstar}
 \theta_t^* = \argmin_{\theta\in\Theta_0}\sum_{i=1}^n \ell_{t,i}(\theta)
\end{equation}
and
\begin{equation}
\label{learning:pi:criterion}
 \mathcal{L}_t(\pi) = \min_{\theta\in\Theta_0} \sum_{i=1}^n \ell_{t,i}(\theta)+ C\log \frac{1}{\pi(\theta)}
\end{equation}
for $\pi=(\pi(\theta_1),\dots,\pi(\theta_M)) \in \Lambda$ with
\begin{equation}
 \Lambda = \left\{ x\in(\mathbb{R}_+)^M\text{: } \sum_{h=1}^M x_h = 1 \text{ and } x_h\geq \frac{1}{2M} \right\}.
\end{equation}
It is direct to check that $\mathcal{L}_t$ is convex and $L$-Lipschitz with $L=2 C M$ on $\Lambda$, this allows to use OPMS (or OGMS).

\begin{thm}
 \label{thm:learning:pi}
 Under Assumption~\ref{asm:loss:Bayes:2}, using OPMS on $\mathcal{L}_t(\eta)$ as in~\ref{learning:pi:criterion} with $\pi_1=(1/M,\dots,1/M)$, $L=2CM$ and
\begin{equation}
 \alpha = \frac{1}{2CM\sqrt{T}},
\end{equation}
 define $I^*=\{\theta_1^*,\dots,\theta_T^*\}$ where each $\theta_t^*$ is as in~\eqref{thetatstar} and $m^*={\rm card}(I^*)$. We have
\begin{equation}
  \sum_{t=1}^T \sum_{i=1}^n \ell_{t,i}(\theta_{t,i}^{\pi_t}) \leq \sum_{t=1}^T   \sum_{i=1}^n \ell_{t,i}(\theta_t^*) + C T\log (2 m^*)
 + 2CM\sqrt{T}.
\end{equation}
\end{thm}

When learning in isolation with a uniform prior, the meta-regret is in $T C \log(M)$. On the other hand, if $m^*$ is small, meta-learning leads to a meta-regret in $CT\log(2m^*) + 2CM\sqrt{2T}$. For $T$ large enough, this is an important improvement.

\subsection{Discussion on the continuous case}

Let us now discuss the possibility of meta-learning for generalized Bayesian methods when $\Theta_0$ is no longer a finite set. There is a general formula for EWA, given by
\begin{equation}
\label{update:Bayes:exact}
 \rho_{t,i}({\rm d}\theta) = \argmin_{\rho} \left\{ \mathbb{E}_{\theta\sim\rho}\left[\sum_{j=1}^{i-1} \ell_{t,j}(\theta) \right] + \frac{\mathcal{K}(\rho,\pi)}{\eta} \right\}
\end{equation}
where the minimum is taken over all probability distributions absolutely continuous with $\pi$,  $\pi$ is a prior distribution, $\eta>0$ a learning rate and $\mathcal{K}$ the Kullback-Leibler divergence (KL). Meta-learning for such an update rule is proven in~\cite{pmlr-v54-alquier17a,MaiCS} but usually does not lead to feasible strategies. Online variational inference~\cite{LinKhan,cherief2019generalization} consists in replacing the minimization on the set of all probability distributions by minimization in a smaller set in order to define a feasible approximation of $\rho_{t,i}$. For example, let $(q_\mu)_{\mu\in M}$ be a parametric family of probability distributions, we define:
\begin{equation}
\label{update:Bayes}
 \mu_{t,i} = \argmin_{\mu\in M} \left\{ \mathbb{E}_{\theta\sim q_\mu}\left[\sum_{j=1}^{i-1} \ell_{t,j}(\theta) \right] + \frac{\mathcal{K}(q_\mu,\pi)}{\eta} \right\}.
\end{equation}
It is discussed in~\cite{ConvexDomke} that generally, when $\mu$ is a location-scale parameter and $\ell_{t,j}$ is $\Gamma$-Lipschitz and convex, then $\bar{\ell}_{t,i}(\mu) :=  \mathbb{E}_{\theta\sim q_\mu}[\ell_{t,j}(\theta)]$ is $2\Gamma$-Lipschitz and convex. In this case, under the assumption that $\mathcal{K}(q_\mu,\pi)$ is $\alpha$-strongly convex in $\mu$, a regret bound for such strategies was derived in~\cite{cherief2019generalization}:
\begin{equation}
\sum_{i=1}^n \mathbb{E}_{\theta\sim q_{\mu_{t,i}}}\left[ \ell_{t,i}(\theta)\right] \leq \inf_{\mu\in \mathcal{M}} \Biggl\{\mathbb{E}_{\theta\sim q_\mu}\left[ \sum_{i=1}^n\ell_{t,i}(\theta)\right]
+ \frac{ \eta 4\Gamma^2 n}{\alpha} + \frac{\mathcal{K}(q_\mu,\pi)}{\eta} \Biggr\}.
\end{equation}
A complete study of meta-learning of the rate $\eta>0$ and of the prior $\pi$ in this context is an important objective (possibly, with a restriction that $\pi\in\{q_\mu,\mu\in M\}$). However, this raises many problems. For example, the KL divergence $\mathcal{K}(q_\mu,q_{\mu'})$ is not always convex with respect to the parameter $\mu'$. In this case it might help to replace it by a convex relaxation that would allow to use OGMS or OPMS. This relates to~\cite{knoblauch2019generalized,alquier2020non} who advocate to go beyond the KL divergence in~\eqref{update:Bayes:exact}, see also~\cite{alquier} and the references therein. This will be the object of future works.

\section{Proofs}
\label{section:proofs}

We start with a preliminary lemma that will be used in the proof of Proposition~\ref{prop:OPMS:OGMS}.

\begin{lemma}
\label{lemma:easy}
 Let $a,b,c$ be three vectors in $\mathbb{R}^p$. Then:
 \begin{equation}
  (a-b)^T(b-c) = \frac{\|a-c\|^2 - \|a-b\|^2 - \|b-c\|^2}{2}.
 \end{equation}
\end{lemma}

\noindent {\it Proof:} expand $\|a-c\|^2 = \|a\|^2 + \|c\|^2 - 2a^T c$ in the r.h.s, as well as $\|a-b\|^2$ and $\|b-c\|^2$. Then simplify. $\square$

We now prove separately Proposition~\ref{prop:OPMS:OGMS} for the general PGMS strategy, and then for OGMS.

\noindent {\it Proof of Proposition~\ref{prop:OPMS:OGMS} for OPMS:} note that up to our knowledge, regret bounds for online updates based on the proximal operator were first studied in Exercice 11.3 in~\cite{cesa2006prediction}. We here provide a detailed proof in our particular setting. Note that better bounds were recently proven in~\cite{campolongo2020temporal}, where the order of the bound is improved under stronger assumptions.

First, $\lambda_t$ is defined as the minimizer of a convex function in~\eqref{dfn:lambda:OGMS}. So, the subdifferential of this function at $\lambda_t$ contains $0$. In other words, there is a $z_{t}\in\partial \mathcal{L}_{t-1}(\lambda_t)$ such that
\begin{equation}
 z_t  =  \frac{\lambda_{t-1} - \lambda_{t}}{\alpha} .
\end{equation}

By convexity, for any $\lambda$, for any $z\in\partial \mathcal{L}_{t-1}(\lambda_t)$,
\begin{equation}
 \mathcal{L}_{t-1}(\lambda)
 \geq \mathcal{L}_{t-1}(\lambda_t) + (\lambda-\lambda_t)^T z.
\end{equation}
The choice $z=z_t$ gives:
\begin{equation}
 \mathcal{L}_{t-1}(\lambda)
 \geq  \mathcal{L}_{t-1}(\lambda_t) + \frac{(\lambda-\lambda_t)^T (\lambda_{t-1} - \lambda_{t})}{\alpha},
\end{equation}
that is,
\begin{align}
 \mathcal{L}_{t-1}(\lambda_t) & \leq  \mathcal{L}_{t-1}(\lambda) + \frac{ (\lambda-\lambda_t)^T (\lambda_{t} - \lambda_{t-1})}{\alpha}
 \nonumber
 \\
 & = \mathcal{L}_{t-1}(\lambda) + \frac{\|\lambda-\lambda_{t-1}\|^2 - \|\lambda-\lambda_{t}\|^2}{2 \alpha}
 -\frac{\|\lambda_t-\lambda_{t-1}\|^2 }{2\alpha}
 \nonumber
 \\
 & = \mathcal{L}_{t-1}(\lambda) + \frac{\|\lambda-\lambda_{t-1}\|^2 - \|\lambda-\lambda_{t}\|^2}{2 \alpha}-\alpha \frac{\|z_t\|^2 }{2}
 \label{equa:proof:prop:OPMS:1}
\end{align}
where we used Lemma~\ref{lemma:easy}. Then, note that
\begin{align}
 \mathcal{L}_{t-1}(\lambda_{t-1})
 & = \mathcal{L}_{t-1}(\lambda_{t}) + [\mathcal{L}_{t-1}(\lambda_{t-1}) - \mathcal{L}_{t-1}(\lambda_{t}) ]
 \nonumber
 \\
 & \leq \mathcal{L}_{t-1}(\lambda_{t}) + \|\lambda_{t-1} - \lambda_t\| L
 \nonumber
 \\
 & \leq \mathcal{L}_{t-1}(\lambda_{t}) + \alpha \| z_t \| L.
\end{align}
Combine this inequality with~\eqref{equa:proof:prop:OPMS:1} gives
\begin{multline}
 \mathcal{L}_{t-1}(\lambda_{t-1}) \leq \mathcal{L}_{t-1}(\lambda)
 +  \frac{ \|\lambda - \lambda_{t-1}\|^2 - \|\lambda - \lambda_{t}\|^2 }{2\alpha}
  \\ + \alpha \left( \|z_t\| L - \frac{\|z_t\|^2}{2} \right).
\end{multline}
Now, for any $x\in\mathbb{R}$, $-x^2/2 + x L - L^2 / 2 \leq 0$. In particular, $\|z_t\| L - \|z_t\|^2/2 \leq L^2/2$ and so the above can be rewritten:
\begin{equation}
 \mathcal{L}_{t-1}(\lambda_{t-1}) \leq \mathcal{L}_{t-1}(\lambda)
 +  \frac{ \|\lambda - \lambda_{t-1}\|^2 - \|\lambda - \lambda_{t}\|^2 }{2\alpha} + \frac{\alpha L^2}{2} .
\end{equation}
Summing the inequality for $t=2$ to $T+1$ leads to:
\begin{equation}
 \sum_{t=1}^T \mathcal{L}_{t}(\lambda_{t}) \leq  \sum_{t=1}^T \mathcal{L}_{t}(\lambda)
 +  \frac{ \|\lambda - \lambda_{1}\|^2 - \|\lambda - \lambda_{T+1}\|^2}{2\alpha} + \frac{\alpha T L^2}{2}.
\end{equation}
This ends the proof. $\square$

\noindent {\it Proof of Proposition~\ref{prop:OPMS:OGMS} for OGMS:} the beginning of the proof follows the proof of Theorem 11.1 in~\cite{cesa2006prediction}.

Note that we can rewrite~\eqref{dfn:lambda:OGMS} as
\begin{equation*}
 \left\{
 \begin{array}{l}
 \tilde{\lambda}_{t} = \lambda_{t-1} - \alpha \nabla \mathcal{L}_{t-1}
(\lambda_{t-1})
\\
\lambda_t = \Pi_{\Lambda}(\tilde{\lambda}_{t})
\end{array}
 \right.
\end{equation*}
Rearranging the first line, we obtain:
\begin{equation}
 \nabla \mathcal{L}_{t-1}(\lambda_{t-1}) = \frac{\lambda_{t-1} - \tilde{\lambda}_t}{\alpha} .
\end{equation}

By convexity, for any $\lambda$,
\begin{align}
 \mathcal{L}_{t-1} (\lambda)
 &
 \geq \mathcal{L}_{t-1}(\lambda_{t-1}) + (\lambda-\lambda_{t-1})^T \nabla \mathcal{L}_{t-1}(\lambda_{t-1})
 \\
 &
 = \mathcal{L}_{t-1}(\lambda_{t-1}) + \frac{(\lambda-\lambda_{t-1})^T (\lambda_{t-1} - \tilde{\lambda}_t)}{\alpha},
\end{align}
that is,
\begin{equation}
\label{equa:proof:prop:OGMS:1}
 \mathcal{L}_{t-1}(\lambda_{t-1}) \leq  \mathcal{L}_{t-1}(\lambda) - \frac{(\lambda-\lambda_{t-1})^T (\lambda_{t-1} - \tilde{\lambda}_t)}{\alpha}.
\end{equation}
Lemma~\ref{lemma:easy} gives:
\begin{align}
 (& \lambda-\lambda_{t-1})^T (\lambda_{t-1} - \tilde{\lambda}_{t})
 \nonumber
 \\
 & = \frac{ \|\lambda - \tilde{\lambda}_t\|^2 - \|\lambda - \lambda_{t-1}\|^2 - \|\lambda_{t-1} - \tilde{\lambda}_{t}\|^2 }{2}
  \nonumber
 \\
 & = \frac{ \|\lambda - \tilde{\lambda}_t\|^2 - \|\lambda - \lambda_{t-1}\|^2 - \alpha^2 \|\nabla \mathcal{L}_{t-1}(\lambda_{t-1})\|^2 }{2}
 \\
 & \geq \frac{ \|\lambda - \lambda_t\|^2 - \|\lambda - \lambda_{t-1}\|^2 - \alpha^2 \|\nabla \mathcal{L}_{t-1}(\lambda_{t-1})\|^2 }{2},
 \label{equa:proof:prop:OGMS:2}
\end{align}
the last step being justified by:
\begin{equation}
 \|\lambda - \tilde{\lambda}_t\|^2 \geq \|\lambda - \Pi_{\Lambda} (\tilde{\lambda}_t) \|^2 =  \|\lambda - \lambda_t\|^2 
\end{equation}
for any $\lambda \in\Lambda$. Plug \eqref{equa:proof:prop:OGMS:2} in \eqref{equa:proof:prop:OGMS:1} to get:
\begin{multline}
 \mathcal{L}_{t-1}(\lambda_{t-1}) \leq \mathcal{L}_{t-1}(\lambda)
 \\
 +  \frac{ \|\lambda - \lambda_{t-1}\|^2 - \|\lambda - \lambda_{t}\|^2}{2\alpha} + \frac{\alpha \|\nabla \mathcal{L}_{t-1}(\lambda_{t-1})\|^2 }{2}
\end{multline}
and the Lipschitz assumption gives:
\begin{equation}
 \mathcal{L}_{t-1}(\lambda_{t-1}) \leq \mathcal{L}_{t-1}(\lambda)
 +  \frac{ \|\lambda - \lambda_{t-1}\|^2 - \|\lambda - \lambda_{t}\|^2}{2\alpha} + \frac{\alpha L^2}{2}
\end{equation}
Sum the inequality for $t=2$ to $T+1$ to get:
\begin{equation}
 \sum_{t=1}^T \mathcal{L}_{t}(\lambda_{t}) \leq  \sum_{t=1}^T \mathcal{L}_{t}(\lambda)
 +  \frac{ \|\lambda - \lambda_{1}\|^2 - \|\lambda - \lambda_{T+1}\|^2}{2\alpha} + \frac{\alpha T L^2}{2}.
\end{equation}
This ends the proof of the statement for OGMS. $\square$

We now provide a lemma that will be useful for the proof of Proposition~\ref{prop:asm:OGA}.

\begin{lemma}
\label{lemma:double:cvx}
 Let $G(u,v)$ be a convex function of $(u,v)\in U\times V$. Define $g(u) = \inf_{v\in V} G(u,v)$. Then $g$ is convex.
\end{lemma}

\noindent {\it Proof:} indeed, let $\lambda\in[0,1]$ and $(x,y)\in U^2$,
\begin{align}
 g( \lambda x & +(1-\lambda) y)
 \\
 & = \inf_{v \in V} G(\lambda x +(1-\lambda) y,v)
 \\
 & \leq G(\lambda x +(1-\lambda) y, \lambda x' + (1-\lambda) y')
 \\
 & \leq \lambda G(x,x') + (1-\lambda) G(y,y')
\end{align}
where the last two inequalities hold for any $(x',y')\in V^2$. Let us now take the infimum with respect to $(x',y')\in V^2$ in both sides, this gives:
\begin{align}
 g( \lambda x & +(1-\lambda) y)
 \\
 & \leq \inf_{x'\in V} \lambda G(x,x') + \inf_{y'\in V}(1-\lambda) G(y,y')
 \\
 & = \lambda g(x) + (1-\lambda)g(y),
\end{align}
that is, $g$ is convex. $\square$

\noindent {\it Proof of Proposition~\ref{prop:asm:OGA}:} apply Lemma~\ref{lemma:double:cvx} to $u=(\vartheta,\gamma)$, $v=\theta$, $U = \Lambda$, $V = \Theta$ and
\begin{equation}
G(u,v) = \sum_{i=1}^n \ell_{i,t}(\theta) + \frac{\gamma \Gamma^2 n}{2} + \frac{\|\vartheta-\theta\|^2}{2\gamma}.
\end{equation}
This shows $g(u) = \mathcal{L}_t((\vartheta,\gamma))$ is convex with respect $(\vartheta,\gamma)$. Also, $G$ is differentiable w.r.t $u=(\vartheta,\gamma)$, so
\begin{equation}
 \frac{\partial G}{\partial \vartheta}  = \frac{\vartheta-\theta}{\gamma}
 \text{, and }
\frac{\partial G}{\partial \gamma}  = \frac{n\Gamma^2}{2} - \frac{\|\vartheta-\theta\|^2}{2\gamma^2}.
\end{equation}
As a consequence, for $(\theta,\vartheta)\in\Theta^2$ and $\underline{\gamma}\leq \gamma \leq \overline{\gamma} $,
\begin{equation}
 \left\|\frac{\partial G}{\partial \vartheta}\right\|^2 \leq \frac{4 C^2}{\underline{\gamma}^2}
 \text{, and}
\left| \frac{\partial G}{\partial \gamma}\right|^2 \leq \frac{n^2\Gamma^4}{4} + \frac{4 C^4}{\underline{\gamma}^4}.
\end{equation}
This leads to
\begin{align}
 \|\nabla_u G(u,v) \| & = \sqrt{\left\|\frac{\partial G}{\partial \vartheta}\right\|^2 + \left| \frac{\partial G}{\partial \gamma}\right|^2}
 \\
 & = \sqrt{ \frac{n^2\Gamma^4}{4} + \frac{4 C^2}{\underline{\gamma}^2} +  \frac{4 C^4}{\underline{\gamma}^4}} =: L,
\end{align}
that is, for each $v$, $G(u,v)$ is $L$-Lipschitz in $u$. So $g(u) = \inf_{v\in V} G(u,v)$ is $L$-Lipschitz in $u$. $\square$

\noindent {\it Proof of Theorem~\ref{thm:main}:} thanks to the Assumption~\ref{asm:loss}, we can apply Proposition~\ref{prop:asm:OGA}. That is, Assumption~\eqref{asm:regret} is satisfied, and we can apply Proposition~\ref{prop:OPMS:OGMS}. This gives:
\begin{multline}
\label{eq:proof:thm:1}
 \sum_{t=1}^T \sum_{i=1}^n \ell_{t,i}(\theta_{t,i}^{\lambda_t})
\leq
 \inf_{\theta_1,\dots,\theta_T \in \Theta} \inf_{(\vartheta,\gamma)\in\Lambda} \Biggl\{
 \sum_{t=1}^T \Biggl[ \sum_{i=1}^n \ell_{t,i}(\theta_{t})
 \\+ \frac{\gamma \Gamma^2 n}{2} + \frac{\|\theta_t-\vartheta\|^2}{2\gamma}
 \Biggr]
+ \frac{\alpha T L^2}{2} + \frac{\|\vartheta-\vartheta_1\|^2 + |\gamma-\gamma_1|^2 }{2\alpha}
\Biggr\}
.
\end{multline}
We use direct bounds for the last two terms: $\|\vartheta-\vartheta_1\|^2 \leq 4C^2$ and $|\gamma-\gamma_1|^2 \leq |\overline{\gamma}-\underline{\gamma}|^2 \leq \overline{\gamma}^2 = C^4 $. Then note that
\begin{align}
 \sum_{t=1}^T  \|\theta_t-\vartheta\|^2 
& = T \left\|\vartheta - \frac{1}{T}\sum_{s=1}^T \theta_s \right\|^2 +
\sum_{t=1}^T \left\|\theta_t - \frac{1}{T}\sum_{s=1}^T \theta_s \right\|^2
\\
& = T \left\|\vartheta - \frac{1}{T}\sum_{s=1}^T \theta_s \right\|^2 + T \sigma^2(\theta_{1}^T).
\end{align}
Upper bounding the infimum on $\vartheta$ in~\eqref{eq:proof:thm:1} by $\vartheta = \frac{1}{T}\sum_{s=1}^T \theta_s$ leads to
\begin{multline}
\label{eq:proof:thm:2}
 \sum_{t=1}^T \sum_{i=1}^n \ell_{t,i}(\theta_{t,i}^{\lambda_t})
\leq
 \inf_{\theta_1,\dots,\theta_T \in \Theta} \inf_{\gamma\in[\underline{\gamma},\overline{\gamma}]} \Biggl\{
 \sum_{t=1}^T  \sum_{i=1}^n \ell_{t,i}(\theta_{t})
+ \frac{\gamma \Gamma^2 n T}{2}
\\ + \frac{T \sigma^2(\theta_{1}^T) }{2\gamma}
+ \frac{\alpha T L^2}{2} + \frac{C^2(4+C^2) }{2\alpha}
\Biggr\}.
\end{multline}
The right-hand side of~\eqref{eq:proof:thm:2} is minimized with respect to $\alpha$ if $\alpha = \frac{C}{L}\sqrt{\frac{4+C^2}{T}}$, which is the value proposed in the theorem, and we obtain:
\begin{multline}
\label{eq:proof:thm:3}
 \sum_{t=1}^T \sum_{i=1}^n \ell_{t,i}(\theta_{t,i}^{\lambda_t})
\leq
 \inf_{\theta_1,\dots,\theta_T \in \Theta} \inf_{\gamma\in[\underline{\gamma},\overline{\gamma}]} \Biggl\{
 \sum_{t=1}^T  \sum_{i=1}^n \ell_{t,i}(\theta_{t})
 \\+ \frac{\gamma \Gamma^2 n T}{2} + \frac{T \sigma^2(\theta_{1}^T) }{2\gamma}
+ CL \sqrt{(4+C^2)T} \Biggr\}.
\end{multline}
The infimum with respect to $\gamma$ in the r.h.s is reached for
\begin{equation}
\gamma^* = \left(\underline{\gamma} \vee \frac{\sigma(\theta_{1}^T)}{\Gamma \sqrt{n}}\right)\wedge \overline{\gamma}.
\end{equation}
First, note that
\begin{align}
\label{eq:proof:thm:4}
 \frac{\gamma^* \Gamma^2 n T}{2}
 & \leq \left(\underline{\gamma} \vee \frac{\sigma(\theta_{1}^T)}{\Gamma \sqrt{n}}\right) \frac{\Gamma^2 n T}{2} 
 \\
 & \leq \left(\underline{\gamma} + \frac{\sigma(\theta_{1}^T)}{\Gamma \sqrt{n}}\right) \frac{ \Gamma^2 n T}{2}
 \\
 & = \frac{\Gamma^2 T n^{1-\beta}}{2} + \frac{\sigma(\theta_{1}^T) \Gamma T \sqrt{n}}{2},
\end{align}
using $\underline{\gamma}=n^{-\beta}$. Then,
\begin{align}
\label{eq:proof:thm:5}
  \frac{T \sigma^2(\theta_{1}^T) }{2\gamma^*}
  & \leq   \frac{T \sigma^2(\theta_{1}^T) }{2} \left( \frac{1}{\overline{\gamma}} \vee \frac{\Gamma \sqrt{n}}{\sigma(\theta_{1}^T)} \right)
  \\
  & \leq \frac{T \sigma^2(\theta_{1}^T) }{2} \left( \frac{1}{\overline{\gamma}} + \frac{\Gamma \sqrt{n}}{\sigma(\theta_{1}^T)} \right)
  \\
  & = \frac{T \sigma^2(\theta_{1}^T) }{2 C^2} + \frac{\sigma(\theta_{1}^T) \Gamma T \sqrt{n}}{2}
  \\
  & \leq \frac{T \sigma(\theta_{1}^T) }{C} + \frac{\sigma(\theta_{1}^T) \Gamma T \sqrt{n}}{2},
\end{align}
using $\overline{\gamma}=C^2$ and $\sigma(\theta_{1}^T) \leq 2C$. Pluging~\eqref{eq:proof:thm:4},~\eqref{eq:proof:thm:5} and the definition of $L$ into~\eqref{eq:proof:thm:3} gives
\begin{multline}
 \sum_{t=1}^T \sum_{i=1}^n \ell_{t,i}(\theta_{t,i}^{\lambda_t})
\leq
 \inf_{\theta_1,\dots,\theta_T \in \Theta} \Biggl\{
 \sum_{t=1}^T  \sum_{i=1}^n \ell_{t,i}(\theta_{t})
 \\
+ C \sqrt{\left(\frac{n^2 \Gamma^4}{4} + 4C^2 n^{2\beta} + 4C^4 n^{4\beta} \right) (4+C^2)T}
 \\
 + \frac{\Gamma^2 T n^{1-\beta}}{2}
  + \sigma(\theta_{1}^T) T \left(\Gamma\sqrt{n} + \frac{1}{C}\right)
 \Biggr\}.
\end{multline}
This ends the proof. $\square$

\noindent {\it Proof of Theorem~\ref{thm:learning:eta}:} a direct application of Proposition~\ref{prop:OPMS:OGMS} gives
\begin{multline}
   \sum_{t=1}^T \sum_{i=1}^n \ell_{t,i}(\theta_{t,i}^{\eta_t}) \leq \inf_{\eta\geq \frac{1}{n}} \Biggl\{ \sum_{t=1}^T  \min_{\theta\in\Theta_0}  \Biggl[ \sum_{i=1}^n \ell_{t,i}(\theta) 
   \\
   +  \frac{\eta n \left[\max_{\vartheta\in\Theta_0,1\leq i\leq n}\ell_{t,i}(\vartheta)\right]^2 }{8} + \frac{\log M}{\eta} \Biggr]
   + \frac{\alpha T L^2}{2} + \frac{(\eta-1)^2}{2\alpha}
   \Biggr\}.
\end{multline}
Thus we have
\begin{multline}
   \sum_{t=1}^T \sum_{i=1}^n \ell_{t,i}(\theta_{t,i}^{\eta_t}) \leq \inf_{\eta\geq \frac{1}{n}} \Biggl\{ \sum_{t=1}^T  \min_{\theta\in\Theta_0}  \Biggl[ \sum_{i=1}^n \ell_{t,i}(\theta) 
   +  \frac{\eta n b^2 }{8} + \frac{\log M}{\eta} \Biggr]
   \\
   + \frac{\alpha T L^2}{2} + \frac{(\eta-1)^2}{2\alpha}
   \Biggr\}.
\end{multline}
Now, pluging in the right-hand side
\begin{equation}
\eta = \frac{1}{n} \vee \left(\frac{2}{b}\sqrt{\frac{2\log M}{n}}\right)\wedge 1,
\end{equation}
we obtain:
\begin{multline}
   \sum_{t=1}^T \sum_{i=1}^n \ell_{t,i}(\theta_{t,i}^{\eta_t}) \leq  \sum_{t=1}^T  \min_{\theta\in\Theta_0}  \Biggl[ \sum_{i=1}^n \ell_{t,i}(\theta) 
   +  \frac{b^2 }{8} + b\sqrt{\frac{n \log(M)}{2}}  +\log(M) \Biggr]
   \\
   + \frac{\alpha T L^2}{2} + \frac{1}{2\alpha}.
\end{multline}
Now, we see that the value $\alpha = \sqrt{2/(TL^2)}$ leads to:
\begin{multline}
   \sum_{t=1}^T \sum_{i=1}^n \ell_{t,i}(\theta_{t,i}^{\eta_t}) \leq  \sum_{t=1}^T  \min_{\theta\in\Theta_0}  \Biggl[ \sum_{i=1}^n \ell_{t,i}(\theta)
   +  \frac{b^2 }{8} + b\sqrt{\frac{n \log(M)}{2}}+\log(M) \Biggr]
   \\
   + L\sqrt{2T} .
\end{multline}
Rearranging terms, and replacing $L$ by its value,
\begin{multline}
   \sum_{t=1}^T \sum_{i=1}^n \ell_{t,i}(\theta_{t,i}^{\eta_t}) \leq  \sum_{t=1}^T  \min_{\theta\in\Theta_0} \sum_{i=1}^n \ell_{t,i}(\theta)
   +  b T \sqrt{\frac{n \log(M)}{2}}
   +  \frac{b^2 T }{8} + T\log(M)
   \\
   + \left(n^2\log M + \frac{nB^2}{8} \right)\sqrt{2T},
\end{multline}
that is the statement of the theorem.
$\square$

\noindent {\it Proof of Theorem~\ref{thm:learning:pi}:} an application of Proposition~\ref{prop:OPMS:OGMS} leads to
\begin{multline}
  \sum_{t=1}^T \sum_{i=1}^n \ell_{t,i}(\theta_{t,i}^{\pi_t}) \leq \min_{\pi\in\Lambda} \Biggl\{\sum_{t=1}^T  \Biggl[\sum_{i=1}^n \ell_{t,i}(\theta_t^*) + C \log \frac{1}{\pi(\theta_t^*)}
 \Biggr]
 \\
 + \frac{\alpha T L^2}{2} + \frac{\|\pi-\pi_1\|^2}{2 \alpha}
 \Biggr\}
\end{multline}
and so
\begin{multline}
  \sum_{t=1}^T \sum_{i=1}^n \ell_{t,i}(\theta_{t,i}^{\pi_t}) \leq \min_{\pi\in\Lambda} \Biggl\{\sum_{t=1}^T  \Biggl[\sum_{i=1}^n \ell_{t,i}(\theta_t^*) + C \log \frac{1}{\pi(\theta_t^*)}
 \Biggr]
 \\
 + \frac{\alpha T L^2}{2} + \frac{1}{2 \alpha}
 \Biggr\}
\end{multline}
Define $\pi_{I^*} $ such that $\pi_{I^*}(\theta_j)=1/(2 m^*)$ if $j\in I^*$ and $\pi_{I^*}(\theta_j)=1/(2 (M-m^*))$ otherwise. We have $\pi_{I}^*\in\Lambda$ and thus
\begin{equation}
  \sum_{t=1}^T \sum_{i=1}^n \ell_{t,i}(\theta_{t,i}^{\pi_t}) \leq \sum_{t=1}^T  \Biggl[\sum_{i=1}^n \ell_{t,i}(\theta_t^*) + C \log (2m^*)
 \Biggr]
 + \frac{\alpha T L^2}{2} + \frac{1}{2 \alpha}.
\end{equation}
Replace $L$ and $\alpha$ by their values to get the theorem. $\square$

\section{Conclusion}

We proposed two simple meta-learning strategies together with their theoretical analysis. Our results clearly show an improvement on learning in isolation if the tasks are similar enough. These theoretical findings are confirmed by our numerical experiments. Important questions remain open. In~\cite{denevi2019online}, a purely online method is proposed, in the sense that it does not require to store all the information of the current task. In the case of OGA, this method allows to learn the starting point. However, its application to learn the step size is not direct~\cite{dimitri}.
An important question is then: is there a purely online method that would
provably improve on learning in isolation in this case? Another important
question is the automatic calibration of $\Gamma$. However, as mentioned in Section~\ref{section:example:Bayes},
we believe that a very general and efficient meta-learning method for learning priors in Bayesian statistics (or in generalized Bayesian inference) would be extremely valuable in practice.

\section*{Acknowledgements}

This project was initiated as Dimitri Meunier's internship project at RIKEN AIP, in the Approximate Bayesian Inference team. The internship was cancelled because of the pandemics. We would like to thank Arnak Dalalyan (ENSAE Paris), who provided fundings so that the internship could take place at ENSAE Paris instead. We would like to thank Emtiyaz Khan (RIKEN AIP), S\'ebastien Gerchinovitz (IRT Saint-Exup\'ery, Toulouse), Vianney Perchet (ENSAE Paris) and all the members of the Approximate Bayesian Inference team for valuable feedback.

\bibliographystyle{abbrv}

\begin{thebibliography}{10}

\bibitem[1]{alquier}
Alquier, P.
\newblock Approximate Bayesian Inference.
\newblock \emph{Entropy}, 22(11):1272, 2020.

\bibitem[2]{alquier2020non}
Alquier, P.
\newblock Non-exponentially weighted aggregation: regret bounds for unbounded
  loss functions.
\newblock In Meila, M. and Zhang, T. (eds.), \emph{Proceedings of the 38th International Conference on Machine Learning},
  volume~139 of \emph{Proceedings of Machine Learning Research}, pp.\ 207--218,
  2021.

\bibitem[3]{pmlr-v54-alquier17a}
Alquier, P., Mai, T.~T., and Pontil, M.
\newblock {Regret Bounds for Lifelong Learning}.
\newblock In Singh, A. and Zhu, J. (eds.), \emph{Proceedings of the 20th
  International Conference on Artificial Intelligence and Statistics},
  volume~54 of \emph{Proceedings of Machine Learning Research}, pp.\  261--269,
  2017.

\bibitem[4]{amit2018meta}
Amit, R. and Meir, R.
\newblock Meta-learning by adjusting priors based on extended {PAC-Bayes}
  theory.
\newblock In \emph{International Conference on Machine Learning}, pp.\
  205--214, 2018.

\bibitem[5]{andrychowicz2016learning}
Andrychowicz, M., Denil, M., Gomez, S., Hoffman, M.~W., Pfau, D., Schaul, T.,
  Shillingford, B., and De~Freitas, N.
\newblock Learning to learn by gradient descent by gradient descent.
\newblock In \emph{Advances in neural information processing systems}, pp.\
  3981--3989, 2016.

\bibitem[6]{balcan2019provable}
Balcan, M.-F., Khodak, M., and Talwalkar, A.
\newblock Provable guarantees for gradient-based meta-learning.
\newblock In \emph{International Conference on Machine Learning}, pp.\
  424--433. PMLR, 2019.

\bibitem[7]{baxter1998theoretical}
Baxter, J.
\newblock Theoretical models of learning to learn.
\newblock In \emph{Learning to learn}, pp.\  71--94. Springer, 1998.

\bibitem[8]{campolongo2020temporal}
Campolongo, N. and Orabona, F.
\newblock Temporal variability in implicit online learning.
\newblock \emph{Advances in Neural Information Processing Systems}, 33, 2020.

\bibitem[9]{cesa2006prediction}
Cesa-Bianchi, N. and Lugosi, G.
\newblock \emph{Prediction, learning, and games}.
\newblock {Cambridge University Press}, 2006.

\bibitem[10]{cherief2019generalization}
Ch\'erief-Abdellatif, B.-E., Alquier, P., and Khan, M.~E.
\newblock A generalization bound for online variational inference.
\newblock \emph{Proceedings of The Eleventh Asian Conference on Machine
  Learning, PMLR}, 101:\penalty0 662--677, 2019.

\bibitem[11]{chollet2019measure}
Chollet, F.
\newblock On the measure of intelligence.
\newblock \emph{arXiv preprint arXiv:1911.01547}, 2019.

\bibitem[12]{denevi2018learning}
Denevi, G., Ciliberto, C., Stamos, D., and Pontil, M.
\newblock Learning to learn around a common mean.
\newblock In \emph{Advances in Neural Information Processing Systems}, pp.\
  10169--10179, 2018.

\bibitem[13]{denevi2019learning}
Denevi, G., Ciliberto, C., Grazzi, R., and Pontil, M.
\newblock Learning-to-learn stochastic gradient descent with biased
  regularization.
\newblock \emph{arXiv preprint arXiv:1903.10399}, 2019.

\bibitem[14]{denevi2019online}
Denevi, G., Stamos, D., Ciliberto, C., and Pontil, M.
\newblock Online-within-online meta-learning.
\newblock In \emph{Advances in Neural Information Processing Systems}, pp.\
  13110--13120, 2019.

\bibitem[15]{denevi2020advantage}
Denevi, G., Pontil, M., and Ciliberto, C.
\newblock The advantage of conditional meta-learning for biased regularization
  and fine tuning.
\newblock \emph{Advances in Neural Information Processing Systems}, 33, 2020.

\bibitem[16]{ConvexDomke}
Domke, J.
\newblock Provable smoothness guarantees for black-box variational inference.
\newblock Preprint arXiv:1901.08431, accepted for ICML 2020, 2019.

\bibitem[17]{fallah2020convergence}
Fallah, A., Mokhtari, A., and Ozdaglar, A.
\newblock On the convergence theory of gradient-based model-agnostic
  meta-learning algorithms.
\newblock In \emph{International Conference on Artificial Intelligence and
  Statistics}, pp.\  1082--1092, 2020.

\bibitem[18]{finn2019online}
Finn, C., Rajeswaran, A., Kakade, S., and Levine, S.
\newblock Online meta-learning.
\newblock \emph{arXiv preprint arXiv:1902.08438}, 2019.

\bibitem[19]{hazan2016oco}
Hazan, E.
\newblock Introduction to online convex optimization.
\newblock \emph{Foundations and Trends{\textregistered} in Optimization},
  2:\penalty0 157--325, 01 2016.

\bibitem[20]{jose2020transfer}
Jose, S.~T., Simeone, O., and Durisi, G.
\newblock Transfer meta-learning: Information-theoretic bounds and information
  meta-risk minimization.
  \newblock \emph{arXiv preprint arXiv:2011.02872}, 2020.

\bibitem[21]{khodak2019adaptive}
Khodak, M., Balcan, M.-F. and Talwalkar, A.
\newblock Adaptive Gradient-Based Meta-Learning Methods.
\newblock In \emph{Advances in Neural Information Processing Systems},
pp.5917--5928, 2019.  
  
\bibitem[22]{knoblauch2019generalized}
Knoblauch, J., Jewson, J., and Damoulas, T.
\newblock Generalized variational inference: Three arguments for deriving new
  posteriors.
\newblock \emph{arXiv preprint arXiv:1904.02063}, 2019.

\bibitem[23]{konobeev2020optimality}
Konobeev, M., Kuzborskij, I., and Szepesv{\'a}ri, C.
\newblock On optimality of meta-learning in fixed-design regression with
  weighted biased regularization.
\newblock \emph{arXiv preprint arXiv:2011.00344}, 2020.

\bibitem[24]{hyperband}
Li, L., Jamieson, K., DeSalvo, G., Rostamizadeh, A. and Talwalkar, A.
\newblock Hyperband: A novel bandit-based approach to hyperparameter optimization.
\newblock \emph{The Journal of Machine Learning Research}, 18(1):6765--6816, 2017.

\bibitem[25]{LinKhan}
Lin, W., Khan, M.~E., and Schmidt, M.
\newblock Fast and simple natural-gradient variational inference with mixture
  of exponential-family approximations.
\newblock In Chaudhuri, K. and Salakhutdinov, R. (eds.), \emph{Proceedings of
  the 36th International Conference on Machine Learning}, volume~97 of
  \emph{Proceedings of Machine Learning Research}, pp.\  3992--4002, Long
  Beach, California, USA, 09--15 Jun 2019. PMLR.
  
\bibitem[26]{MaiCS}
Mai, T. T.
\newblock On continual single index learning.
\newblock \emph{arXiv preprint arXiv:2102.12961}, 2021.

\bibitem[27]{maurer2006}
Maurer, A.
\newblock Bounds for linear multi-task learning.
\newblock \emph{Journal of Machine Learning Research}, 7:\penalty0 117--139,
  2006.

\bibitem[28]{MPR2016}
Maurer, A., Pontil, M., and Romera-Paredes, B.
\newblock The benefit of multitask representation learning.
\newblock \emph{Journal of Machine Learning Research}, 17(81):\penalty0 1--32,
  2016.
  
\bibitem[29]{dimitri}
Meunier, D.
\newblock Meta-learning meets variational inference: Learning priors with
  guarantees.
\newblock MSc thesis, Universit\'e Paris Saclay, 2020.
\newblock \url{https://dimitri-meunier.github.io/files/RikenReport.pdf}.

\bibitem[30]{nesterov2013introductory}
Nesterov, Y.
\newblock \emph{Introductory lectures on convex optimization: A basic course},
  volume~87.
\newblock Springer Science \& Business Media, 2004.

\bibitem[31]{orabona2019modern}
Orabona, F.
\newblock A modern introduction to online learning.
\newblock \emph{arXiv preprint arXiv:1912.13213}, 2019.

\bibitem[32]{parikh2014proximal}
Parikh, N. and Boyd, S.
\newblock Proximal algorithms.
\newblock \emph{Foundations and Trends{\textregistered} in optimization},
  1\penalty0 (3):\penalty0 127--239, 2014.

\bibitem[33]{pentina2014pac}
Pentina, A. and Lampert, C.
\newblock A {PAC-Bayesian} bound for lifelong learning.
\newblock In \emph{Proc. 31st International Conference on Machine Learning},
  pp.\  991--999, 2014.

\bibitem[34]{romera2013multilinear}
Romera-Paredes, B., Aung, H., Bianchi-Berthouze, ., and Pontil, M.
\newblock Multilinear multitask learning.
\newblock In \emph{International Conference on Machine Learning}, pp.\
  1444--1452, 2013.

\bibitem[35]{rothfuss2020pacoh}
Rothfuss, J., Fortuin, V., and Krause, A.
\newblock Pacoh: Bayes-optimal meta-learning with pac-guarantees.
\newblock \emph{arXiv preprint arXiv:2002.05551}, 2020.

\bibitem[36]{ruvolo2013ella}
Ruvolo, P. and Eaton, E.
\newblock Ella: An efficient lifelong learning algorithm.
\newblock In \emph{Proc. 30th International Conference on Machine Learning},
  pp.\  507--515. 2013.

\bibitem[37]{shalev2012online}
Shalev-Shwartz, S.
\newblock Online learning and online convex optimization.
\newblock \emph{Foundations and Trends{\textregistered} in Machine Learning},
  4\penalty0 (2):\penalty0 107--194, 2012.
  
\bibitem[38]{shang}
Shang X., Kaufmann E. and Valko M.
\newblock A simple dynamic bandit algorithm for hyper-parameter tuning.
\newblock \emph{6th ICML Workshop on Automated Machine Learning}, 2019.

\bibitem[39]{thrun2012learning}
Thrun, S. and Pratt, L.
\newblock \emph{Learning to learn}.
\newblock Kluwer Academic Publishers, 1998.

\bibitem[40]{yamada2017localized}
Yamada, M., Koh, T., Iwata, T., Shawe-Taylor, J., and Kaski, S.
\newblock Localized lasso for high-dimensional regression.
\newblock In \emph{Artificial Intelligence and Statistics}, pp.\  325--333.
  PMLR, 2017.

\bibitem[41]{zhou2019efficient}
Zhou, P., Yuan, X., Xu, H., Yan, S., and Feng, J.
\newblock Efficient meta learning via minibatch proximal update.
\newblock In \emph{Advances in Neural Information Processing Systems}, pp.\
1534--1544, 2019.

\end{thebibliography}

\end{document}